\documentclass[10pt,journal,final]{IEEEtran}
\ifCLASSINFOpdf
  \usepackage[pdftex]{graphicx}
  \usepackage{color}
  \usepackage{multirow}
  \usepackage{subfigure}
  \usepackage[graphicx]{realboxes}
  \usepackage{booktabs}
\else

\fi
\usepackage{gensymb}
\usepackage{amsfonts,amssymb}
\usepackage{amsmath}
\usepackage{url}

\usepackage{algorithm}
\usepackage{algorithmicx}

\ifCLASSOPTIONcompsoc
  \usepackage[caption=false,font=normalsize,labelfont=sf,textfont=sf]{subfig}
\else
  \usepackage[caption=false,font=footnotesize]{subfig}
\fi
 \usepackage{cite}
 \usepackage{dblfloatfix}
\begin{document}
%
\title{DiffUCD:Unsupervised Hyperspectral Image Change Detection with Semantic Correlation Diffusion Model}
%
%
\author{Xiangrong~Zhang,~\IEEEmembership{Senior~Member,~IEEE}, ~Shunli~Tian, ~Guanchun~Wang, ~Huiyu Zhou, and ~Licheng~Jiao,~\IEEEmembership{Fellow,~IEEE}\\
\thanks{Xiangrong Zhang, Shunli Tian, Guanchun Wang, and Licheng Jiao are with the Key Laboratory of Intelligent Perception and Image Understanding of Ministry of Education, Xidian University, Xi’an, Shaanxi Province 710071, China.}
\thanks{Huiyu Zhou is with the School of Informatics, University of Leicester, Leicester LE1 7RH.U.K.}

\thanks{This work was supported in part by the National Natural Science Foundation of China under Grant 61871306, Grant 62171332.}}%

%
%

\markboth{}%
{Shell \MakeLowercase{\textit{et al.}}: Bare Demo of IEEEtran.cls for IEEE Journals}
%



\maketitle
\begin{abstract}
Hyperspectral image change detection (HSI-CD) has emerged as a crucial research area in remote sensing due to its ability to detect subtle changes on the earth's surface. Recently, diffusional denoising probabilistic models (DDPM) have demonstrated remarkable performance in the generative domain. Apart from their image generation capability, the denoising process in diffusion models can comprehensively account for the semantic correlation of spectral-spatial features in HSI, resulting in the retrieval of semantically relevant features in the original image. In this work, we extend the diffusion model's application to the HSI-CD field and propose a novel unsupervised HSI-CD with semantic correlation diffusion model (DiffUCD). Specifically, the semantic correlation diffusion model (SCDM) leverages abundant unlabeled samples and fully accounts for the semantic correlation of spectral-spatial features, which mitigates pseudo change between multi-temporal images arising from inconsistent imaging conditions. Besides, objects with the same semantic concept at the same spatial location may exhibit inconsistent spectral signatures at different times, resulting in pseudo change. To address this problem, we propose a cross-temporal contrastive learning (CTCL) mechanism that aligns the spectral feature representations of unchanged samples. By doing so, the spectral difference invariant features caused by environmental changes can be obtained. Experiments conducted on three publicly available datasets demonstrate that the proposed method outperforms the other state-of-the-art unsupervised methods in terms of Overall Accuracy (OA), Kappa Coefficient (KC), and F1 scores, achieving improvements of approximately 3.95$\%$, 8.13$\%$, and 4.45$\%$, respectively. Notably, our method can achieve comparable results to those fully supervised methods requiring numerous annotated samples.
\end{abstract}


%
\IEEEpeerreviewmaketitle
\section{Introduction}
%
%
%
%

Change detection (CD) involves using remote sensing technologies to compare and analyze images taken at different times in the same area, detecting changes in ground objects between two or more images \cite{lu2004change}. Hyperspectral data provides continuous spectral information, making it ideal for detecting subtle changes on the Earth's surface. As such, hyperspectral image change detection (HSI-CD) has become an important research focus in remote sensing \cite{liu2019review}, with applications in land use and land cover change \cite{aslami2018object}, ecosystem monitoring, natural disaster damage assessment \cite{rumpf2010early}, and more.

Broadly speaking, HSI-CD can be achieved using supervised and unsupervised method. Most current methods rely on supervised deep learning networks trained with high-quality labeled samples \cite{wang2022spectral,dong2023abundance}. However, obtaining high-quality labeled training samples is costly and time-consuming. Thus, reducing or eliminating the reliance on labeled data is critical to addressing the challenge of HSI-CD.

Although deep learning based supervised HSI-CD methods have shown promising results, they still face several challenges: 1) There are often insufficient labeled samples for HSI-CD, necessitating the need to effectively leverage labeled and unlabeled data to train deep learning networks. 2) HSI-CD involves spatiotemporal data, where changes occur over time and exhibit spatial correlations. While existing approaches primarily focus on extracting features, they often overlook the importance of considering spectral-spatial semantic correlations. Incorporating such correlations is essential for accurate CD. 3) Objects with the same semantic concept at the same spatial location can exhibit different spectral features at different times due to changes in imaging conditions and environments (i.e., the same objects with different spectra). While most deep learning-based CD methods focus on fully extracting spectral features, none of them has investigated extracting spectral difference invariant features caused by environmental changes.

Recently, many unsupervised HSI-CD methods \cite{chakraborty2021unsupervised,lei2021spectral} have been proposed. Unlike supervised methods, unsupervised methods do not require pre-labeled data and can learn features of changed regions using only two HSIs. This confers a significant advantage over supervised methods, as it avoids the need for labor-intensive and time-consuming labeling and mitigates issues such as inaccurate and inconsistent labeling. However, the accuracy of unsupervised methods is often lower than that of supervised methods, despite their ability to function without any annotation information.

Diffusion models have recently demonstrated remarkable successes in image generation and synthesis \cite{kim2022diffusionclip,saharia2022image}. Thanks to their excellent generative capabilities, researchers have begun exploring the application of diffusion models in visual understanding tasks such as semantic segmentation \cite{tan2022semantic,wu2022medsegdiff}, object detection \cite{chen2022diffusiondet}, image colorization \cite{song2020score}, super-resolution \cite{saharia2022image,li2022srdiff}, and more. However, their potential for HSI-CD remains largely unexplored. As such, how to apply diffusion models to HSI-CD remains an open problem.

To address the challenges faced by HSI-CD, we propose an unsupervised approach based on semantic correlation diffusion model (SCDM) that leverages its strong denoising generation ability. This method consists of two main steps. Firstly, the denoising process of the SCDM can utilize many unlabeled samples, fully consider the semantic correlation of spectral-spatial features, and retrieve the features of the original image semantic correlation. Secondly, we propose a cross-temporal contrastive learning (CTCL) mechanism to address the problem of spectral variations caused by environmental changes. This method aligns the spectral feature representations of unchanged samples cross-temporally, enabling the network to learn features that are invariant to these spectral differences.

The main contributions of this paper are:

\begin{itemize}
\item We propose DiffUCD, the first diffusion model designed explicitly for HSI-CD, which can fully consider the semantic correlation of spectral-spatial features and retrieve semantically related features in the original image.

\item To address the problem that objects with the same semantic concept at the same spatial location may exhibit different spectral features at different times, we propose CTCL, which enables the network to learn the spectral difference invariant features. 

\item Extensive experiments on three datasets demonstrate that our proposed method achieves state-of-the-art results compared to other unsupervised HSI-CD methods.

\end{itemize}

Through experiments on three publicly available datasets (Santa Barbara, Bay Area, and Hermiston), we demonstrate that DiffUCD outperforms state-of-the-art methods by a significant margin. Specifically, our method achieves OA values of 96.87$\%$, 96.35$\%$, and 95.47$\%$ on the three datasets, respectively, which are 5.73$\%$, 5.56$\%$, and 0.57$\%$ higher than those achieved by the state-of-the-art unsupervised method. Even when trained with the same number of human-labeled training samples, our method exhibits competitive performance compared to supervised methods. When compared to ML-EDAN \cite{qu2021multilevel}, our method achieves slightly better or similar performance, with OA values changing by $-$1.13$\%$, $-$0.12$\%$, and $+$0.89$\%$, respectively. In summary, our approach extends the application of diffusion models to HSI-CD, achieving superior results compared to previous methods.

The rest of this article is organized as follows. Section II introduces the related work of this paper. Section III introduces the proposed framework for HSI-CD in detail. Section IV introduces the experiments. Finally, the conclusion of this paper is drawn in Section V.

\section{Related Work}

\subsection{Unsupervised HSI-CD}

There has been a growing interest in unsupervised HSI-CD methods based on deep learning in recent years. Recent studies have focused on mitigating the impact of noisy labels in pseudo-labels \cite{zhao2022efficient,li2019unsupervised}. Li et al. \cite{li2019unsupervised} proposed an unsupervised fully convolutional HSI-CD framework based on noise modeling. This framework uses parallel Siamese fully convolutional networks (FCNs) to extract features from bitemporal images separately. The unsupervised noise modeling module can alleviate the accuracy limitation caused by pseudo-labels. An unsupervised method \cite{li2021unsupervised} that self-generates trusted labels has been proposed to improve pseudo-labels' quality. This method combines two model-driven methods, CVA and SSIM, to generate trusted pseudo-training sets, and the trusted pseudo-labels can improve the performance of deep learning networks. While recent advances in unsupervised HSI-CD methods have shown promise, the efficient extraction of changing features remains challenging \cite{hou2021three,liu2023unsupervised}. UTBANet \cite{liu2023unsupervised} aims to reconstruct HSIs and adds a decoding branch to reconstruct edge information. Unlike previous methods, this paper utilizes many unlabeled HSI-CD samples to train SCDM to extract semantically relevant spectral-spatial information.

\subsection{Diffusion models}
Diffusion models \cite{song2020score, ho2020denoising, song2019generative} are Markov chains that reconstruct data samples through a step-by-step denoising process, beginning with randomly distributed samples. Recently, methods based on diffusion models have been brilliant in various fields, such as computer vision \cite{saharia2022image, rombach2022high, brempong2022denoising, wyatt2022anoddpm}, natural language processing \cite{austin2021structured, li2022diffusion}, multimodal learning \cite{avrahami2022blended, yang2023diffsound}, time series modeling \cite{tashiro2021csdi, rasul2021autoregressive}, etc. Diffusion models have been gradually explored in terms of visual representation, and Baranchuk et al. \cite{baranchuk2021label} demonstrated that diffusion models could also be used as a tool for semantic segmentation, especially when labeled data is scarce. Gu et al. \cite{gu2022diffusioninst} proposed a new framework, DiffusionInst, which represents instances as instance-aware filters and instance segmentation as a noise-to-filter denoising process. In this paper, we propose SCDM and further explore the application of the diffusion model in the field of HSI-CD. To our knowledge, this is the first work that employs a diffusion model for HSI-CD.

\begin{figure*}[htbp]
  \centering
  \includegraphics[scale=0.95]{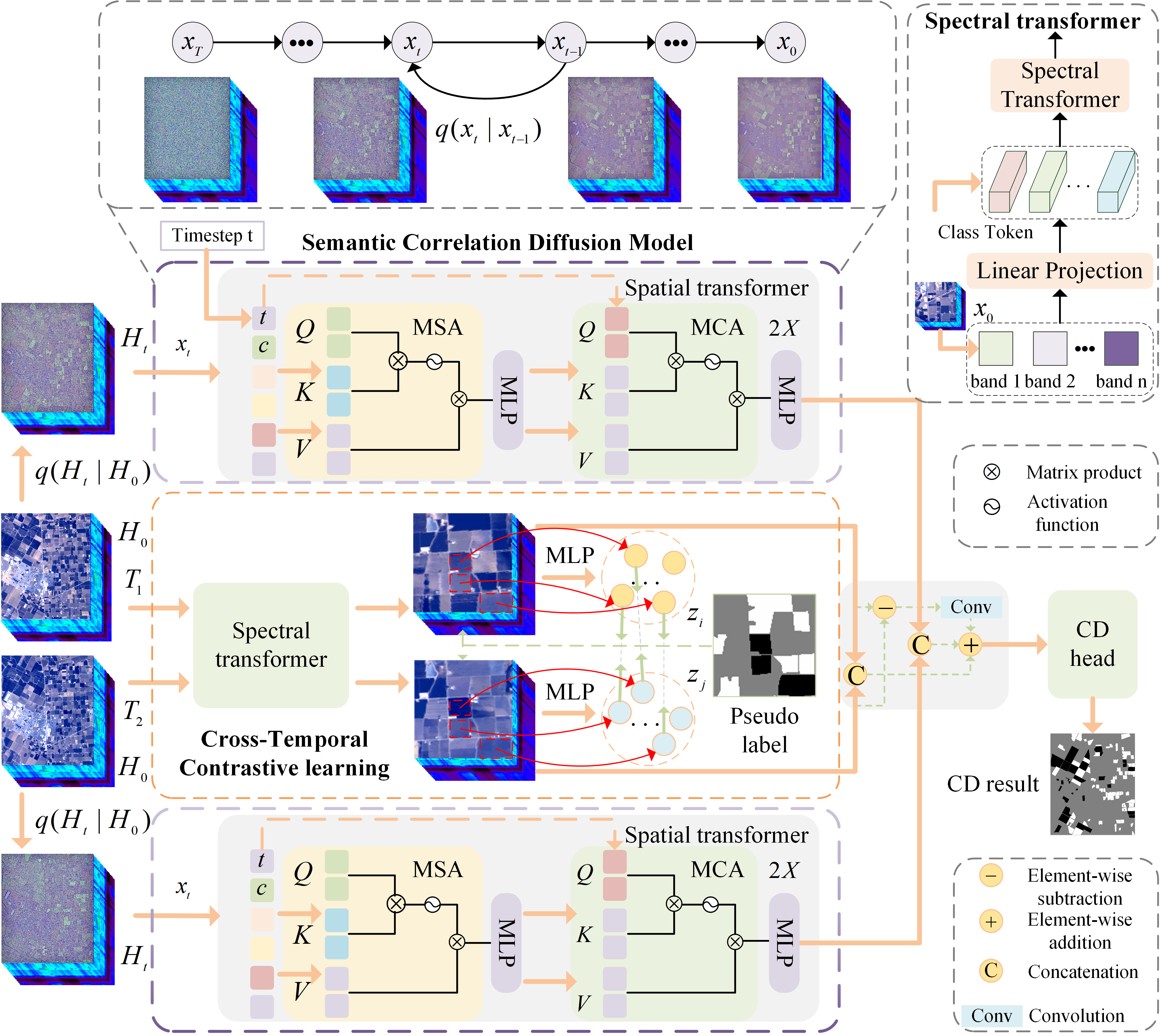}
  \caption{\textbf{The proposed DiffUCD framework} consists of two main modules: SCDM and CTCL. SCDM can fully consider the semantic correlation of spectral-spatial features and reconstruct the essential features of the original image semantic correlation. CTCL can deal with the problem of the same object with different spectra and constrain the network to learn the invariant characteristics of spectral differences caused by environmental changes. }
  \label{fig:overall}
\end{figure*}

\subsection{Contrastive learning}

Contrastive learning \cite{he2020momentum, chen2020big, NDI} learns feature representations of samples by automatically constructing similar and dissimilar samples. BYOL \cite{grill2020bootstrap} relies on the interaction of the online and target networks for learning. An online network is trained from augmented views of an image to predict target network representations of the same image under different augmented views. SimSiam \cite{chen2021exploring} theoretically explained that the essence of twin network representation learning with stop-gradient is the Expectation-Maximization (EM) algorithm. BYOL \cite{grill2020bootstrap} and SimSiam \cite{chen2021exploring} still work without negative samples. Recently, contrastive learning has achieved promising results in HSI classification tasks \cite{hu2021contrastive,guan2022cross}. Ou et al. \cite{ou2022hyperspectral} proposed an HSI-CD framework based on a self-supervised contrastive learning pre-training model and designed a data augmentation strategy based on Gaussian noise for constructing positive and negative samples. In this paper, we design a CTCL network that can extract the invariant features of spectral differences caused by environmental changes, thereby reducing the impact of imaging conditions and environmental changes on CD results.

\section{Proposed Method}
This section will provide an overview of the DDPM framework \cite{ho2020denoising, nichol2021improved, song2020denoising} and describe the proposed DiffUCD model in detail. Fig. \ref{fig:overall} illustrates the architecture of the DiffUCD model, which comprises three main parts: the SCDM, CTCL, and CD head.

\subsection{Preliminaries}

Inspired by nonequilibrium thermodynamics \cite{sohl2015deep}, a series of probabilistic generative models called diffusion models have been proposed. There are currently three popular formulations based on diffusion models: denoising diffusion probabilistic models (DDPMs) \cite{ho2020denoising, nichol2021improved, song2020denoising}, score-based generative models (SGMs) \cite{song2019generative, song2020improved}, and stochastic differential equations (Score SDEs) \cite{song2020score, song2021maximum}. In this paper, we expand the application of DDPMs to the HSI-CD domain.

Diffusion probabilistic models for denoising typically use two Markov chains: a forward chain that perturbs the image with noise and a reverse chain that denoises the noisy image. The forward chain is a process of forward diffusion, which gradually adds Gaussian noise to the input data to create interference. The reverse chain learns a denoising network that reverses the forward diffusion process. In the forward diffusion process of noise injection, Gaussian noise is gradually added to the clean data $x_{0} \sim p\left(x_{0}\right)$ until the data is entirely degraded, resulting in a Gaussian distribution $\mathcal{N}(\mathbf{0}, \mathbf{I})$. Formally, the operation at each time step $t$ in the forward diffusion process is defined as:

\begin{equation}
  q\left(x_{t} \mid x_{t-1}\right)=\mathcal{N}\left(x_{t} ; \sqrt{1-\beta_{t}} x_{t-1}, \beta_{t} \mathbf{I}\right)
  \label{eq:forward}
\end{equation}
Here $\left(x_{0}, x_{1}, \cdots, x_{T}\right)$ represents a $T$-step Markov chain. $\beta_{t} \in(0,1)$ represent the noise Schedule.

Importantly, given a clean data sample $x_{0}$, we can obtain a noisy sample $x_{t}$ by sampling the Gaussian vector $\boldsymbol{\epsilon} \sim \mathcal{N}(\mathbf{0}, \mathbf{I})$ and applying the transformation directly to $x_{0}$:

\begin{equation}
    q\left(x_{t} \mid x_{0}\right)=\mathcal{N}\left(x_{t} \mid x_{0} \sqrt{\bar{\alpha}_{t}},\left(1-\bar{\alpha}_{t}\right) \mathbf{I}\right)
    \label{eq:2}
\end{equation}

\begin{equation}
    x_{t}=x_{0} \sqrt{\bar{\alpha}_{t}}+\epsilon_{t} \sqrt{1-\bar{\alpha}_{t}}, \quad \epsilon_{t} \sim \mathcal{N}(0, \mathbf{I})
    \label{eq:3}
\end{equation}

To add noise to $x_{0}$, we use \text{Eq.} \ref{eq:3} to transform the data into $x_{t}$ for each time step $t \in\{0,1, \ldots, T\}$. Here $\bar{\alpha}_{t}=\prod_{i=0}^{t} \alpha_{i}=\prod_{i=0}^{t}\left(1-\beta_{i}\right)$.

During the training phase, a U-ViT\cite{bao2022all} like structure for $\epsilon_{\theta}\left(x_{t}, t\right)$ is trained to predict $\epsilon$ by minimizing the training objective using L2 loss.

\begin{equation}
    \mathcal{L}=\left\|\epsilon-\epsilon_{\theta}\left(x_{t}, t\right)\right\|^{2} =\left\|\epsilon-\epsilon_{\theta}\left(\sqrt{\overline{\alpha_{t}}} x_{t-1}+\sqrt{1-\overline{\alpha_{t}}} \epsilon, t\right)\right\|^{2}
    \label{eq:4}
\end{equation}

During the inference stage, given a noisy input $x_{t}$, the trained model $\epsilon_{\theta}\left(x_{t}, t\right)$ is used to denoise and obtain $x_{t-1}$. This process can be mathematically represented as follows:

\begin{equation}
    x_{t-1}=\frac{1}{\sqrt{\alpha_{t}}}\left(x_{t}-\frac{1-\alpha_{t}}{\sqrt{1-\bar{\alpha}_{t}}} \epsilon_{\theta}\left(x_{t}, t\right)\right)+\sigma_{t} z
    \label{eq:5}
\end{equation}
where $z \sim \mathcal{N}(\mathbf{0}, \mathbf{I})$ and $\sigma_{t}=\frac{1-\bar{\alpha}_{t-1}}{1-\bar{\alpha}_{t}} \beta_{t}$. $x_{t}$ obtains $x_{0}$ through continuous iteration, $i.e.$, $\boldsymbol{x}_{t} \rightarrow \boldsymbol{x}_{t-1} \rightarrow \boldsymbol{x}_{t-2} \rightarrow  \ldots \rightarrow x_{0}$.

In this work, we aim to address the task of unsupervised HSI-CD using a diffusion model. Specifically, we consider data sample $x_{0}$ as a patch from the HSI at either $T1$ or $T2$. We begin by corrupting $x_{0}$ with Gaussian noise using \text{Eq.} \ref{eq:3} to obtain the noisy input $x_{t}$ for the noise predictor $\epsilon_{\theta}\left(x_{t}, t, c\right)$. We define $\epsilon_{\theta}\left(x_{t}, t, c\right)$ as a noise predictor that can extract spectral-spatial features that are useful for downstream HSI-CD tasks.

\subsection{DiffUCD}

The proposed DiffUCD framework comprises a SCDM, a CTCL, and a CD head, as illustrated in Fig. \ref{fig:overall}. SCDM can use a large number of unlabeled samples to fully consider the semantic correlation of spectral-spatial features and retrieve the features of the original image semantic correlation. CTCL aligns the spectral sequence information of unchanged pixels, guiding the network to extract features that are insensitive to spectral differences resulting from variations in imaging conditions and environments.



\subsubsection{Semantic Correlation Diffusion Model}

We utilize the forward diffusion process proposed by SCDM \cite{ho2020denoising} in \text{Eq.} (\ref{eq:h0}), which corrupts the input HSI $H_{0}$ to obtain $H_{t}$ at a random time step $t$. Fig. \ref{fig:overall} illustrates that the SCDM takes a patch $x_{t} \in R^{C \times K \times K}$ from the $H_{t}$ at time $T1$ or $T2$ as input. Our SCDM is structured similarly to U-ViT \cite{bao2022all}, with the time step $t$, condition $c$, and noise image $x_{t}$ all used as tokens for input into the SCDM. In contrast to the U-ViT long skip connections method, we employ a multi-head cross-attention (MCA) approach for feature fusion between the shallow and deep layers. The noise image $x_{t}$ is fed into $\epsilon_{\theta}\left(x_{t}, t, c\right)$, parameterized by the SCDM. The pixel-level representation $\hat{x}_{0}$ of $x_{0}$ is obtained through the $\epsilon_{\theta}\left(x_{t}, t, c\right)$ network, and the corresponding formula is given as follows:

\begin{equation}
    H_{t}(H_{0}, \epsilon_{t})=H_{0} \sqrt{\bar{\alpha}_{t}}+\epsilon_{t} \sqrt{1-\bar{\alpha}_{t}}
    \label{eq:h0}
\end{equation}
where $\bar{\alpha}_{t}=\prod_{i=0}^{t} \alpha_{i}=\prod_{i=0}^{t}\left(1-\beta_{i}\right)$, $\epsilon_{t} \sim \mathcal{N}(0, \mathbf{I}) $.

\begin{equation}
    \hat{x}_{0}=\frac{1}{\sqrt{\bar{\alpha}_{t}}}\left(x_{t}-\sqrt{1-\bar{\alpha}_{t}} \epsilon_{\theta}\left(x_{t}, t, c\right)\right)
    \label{eq:6}
\end{equation}

\subsubsection{Cross-Temporal Contrastive Learning}

The proposed CTCL module aims to learn more discriminative features for HSI-CD by emphasizing spectral difference invariant features between unchanged samples at $T1$ and $T2$ moments. The architecture consists of two parts: a spectral transformer encoder and an MLP. To construct positive and negative sample pairs, unchanged pixels at the same location but different phases are used as positive samples, while the rest are negative samples. The CTCL network takes $X_{1}$ and $X_{2}$ as input and produces contrastive feature representations $z^{i}$ and $z^{j}$, which are then aligned through a contrastive loss function. This architecture aims to shorten the distance between the feature representations of unchanged pixel samples in different phases, which helps the network extract more robust and invariant features that are less affected by environmental changes.

\subsection{Change Detection Head}
We employ a fusion module to fuse the semantic correlation of spectral-spatial features obtained by the SCDM with the spectral difference-invariant features extracted by CTCL. The module is formulated as follows:



\begin{equation}\label{eq:7}
    \begin{split}
        \hat{X} &= 1/3(Conv(Sub(\hat{X}_{1}, \hat{X}_{2}))\\&+Concat(\hat{X}_{1}, \hat{X}_{2}) + Concat(\hat{x}_{0}^{1}, \hat{x}_{0}^{2}))
    \end{split}
\end{equation}

Here, $\hat{X}_{1}$ and $\hat{X}_{2}$ represent the encoder output features obtained through CTCL, while $\hat{x}_{0}^{1}$ and $\hat{x}_{0}^{2}$ denote the spectral-spatial features extracted by the SCDM. The $Concat (\cdot)$ function is used to superimpose features along the channel dimension, while ${Sub}(\cdot)$ calculates the features' differences. The resulting fused features, $\hat{X}$, are then passed to the CD head to generate the final CD map. The structure of the CD head used in this paper is consistent with the spatial transformer in Fig. \ref{fig:overall}.

\subsection{Training}

The training process comprises two stages: 1) The SCDM is pre-trained using a large number of unlabeled HSI-CD samples to fully consider the semantic correlation of spectral-spatial features and retrieve the features of the original image semantic correlation. 2) A small set of pseudo-label samples are used to train the CTCL network. The spectral-spatial features extracted by the SCDM are fused with the spectrally invariant features learned by the CTCL network and then passed through the CD head to generate the ultimate CD map.

\subsubsection{Pretrained Semantic Correlation Diffusion Model}

To pre-train the SCDM, we selected the Santa Barbara, Bay Area, and Hermiston datasets\footnote{https://citius.usc.es/investigacion/datasets/hyperspectral-change-detection.}, which contain large amounts of unlabeled data. For the input $x_{0}$, we randomly initialized the time $t$ and added noise using \text{Eq.} (\ref{eq:3}) to obtain $x_{t}$. The pre-trained SCDM predicted $x_{t}$ and then calculated the estimated features of the input data $x_{0}$ using \text{Eq.} (\ref{eq:6}). The noise loss for the SCDM is defined as follows:

\begin{equation}\label{eq:8}
    \begin{aligned} \mathcal{L}_{noise} & =\mathbb{E}_{t, \boldsymbol{x}_{0}, c, \boldsymbol{\epsilon}}\sum_{i=1}^{N}\left\|\boldsymbol{\epsilon}^{i}-\epsilon_{\theta}\left(x_{t}^{i}, t, c\right)\right\|^{2} \\& =\mathbb{E}_{t, \boldsymbol{x}_{0}, c, \boldsymbol{\epsilon}}\sum_{i=1}^{N}\left\|\boldsymbol{\epsilon}^{i}-\epsilon_{\theta}\left(\sqrt{\bar{\alpha}_{t}} x_{t}^{i}+\sqrt{1-\bar{\alpha}_{t}} \boldsymbol{\epsilon}, t\right)\right\|^{2}\end{aligned}
\end{equation}
where $\boldsymbol{\epsilon}^{i}$ represents the noise added to the $i$-th sample using \text{Eq.} (\ref{eq:3}), $N$ represents the number of samples.

\begin{figure*}[htbp]
\centering
\subfigure[DSFA\cite{8824216}]
{
    \begin{minipage}[b]{.17\linewidth}
        \centering
        \includegraphics[scale=0.82]{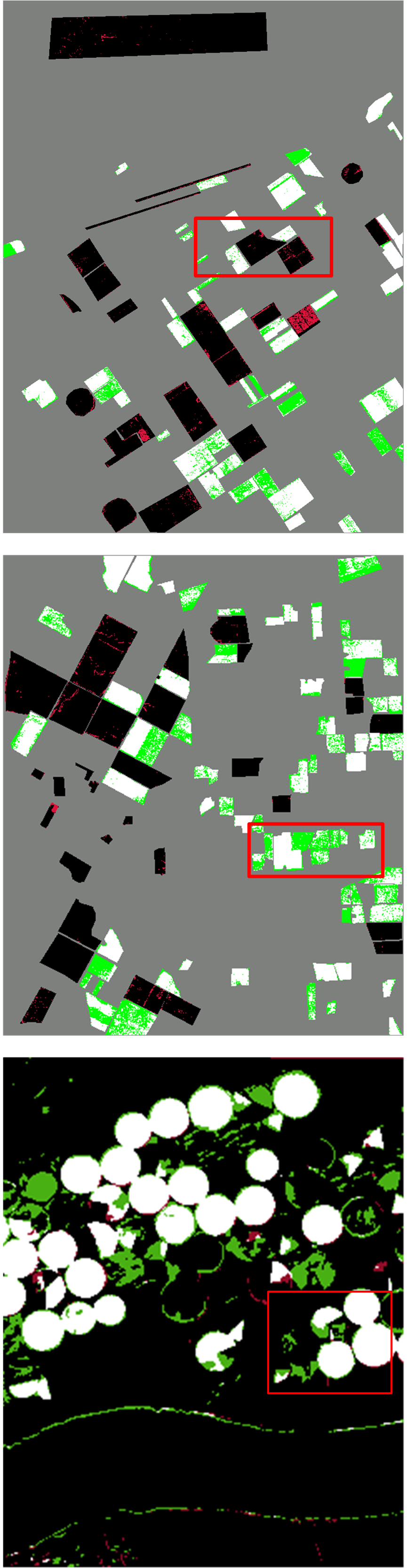}
    \end{minipage}
}%
\subfigure[HyperNet\cite{hu2022hypernet}]
{
 	\begin{minipage}[b]{.17\linewidth}
        \centering
        \includegraphics[scale=0.82]{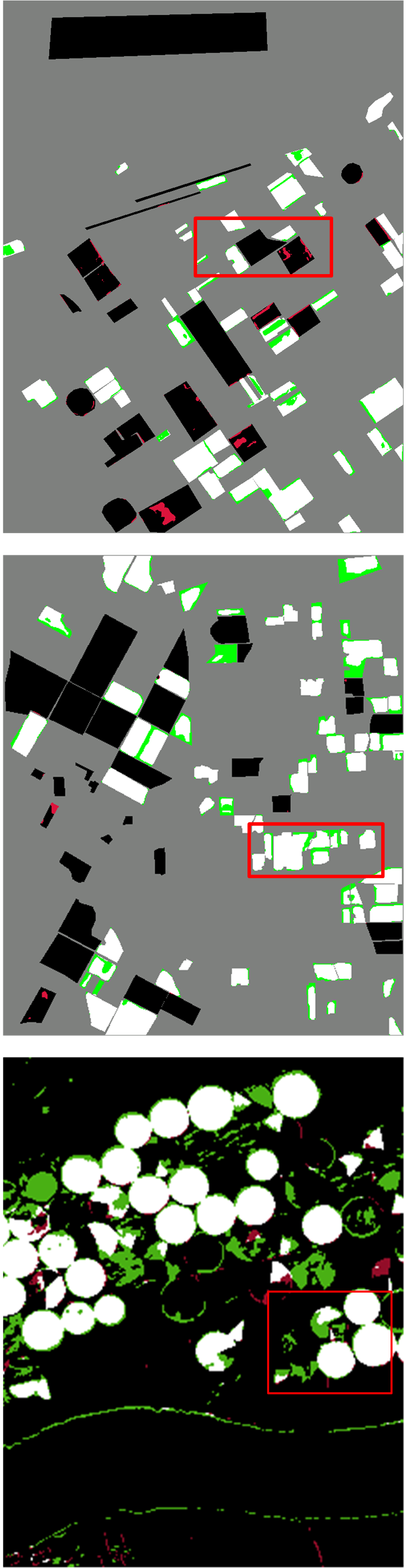}
    \end{minipage}
}
\subfigure[Base+SCDM]
{
 	\begin{minipage}[b]{.17\linewidth}
        \centering
        \includegraphics[scale=0.82]{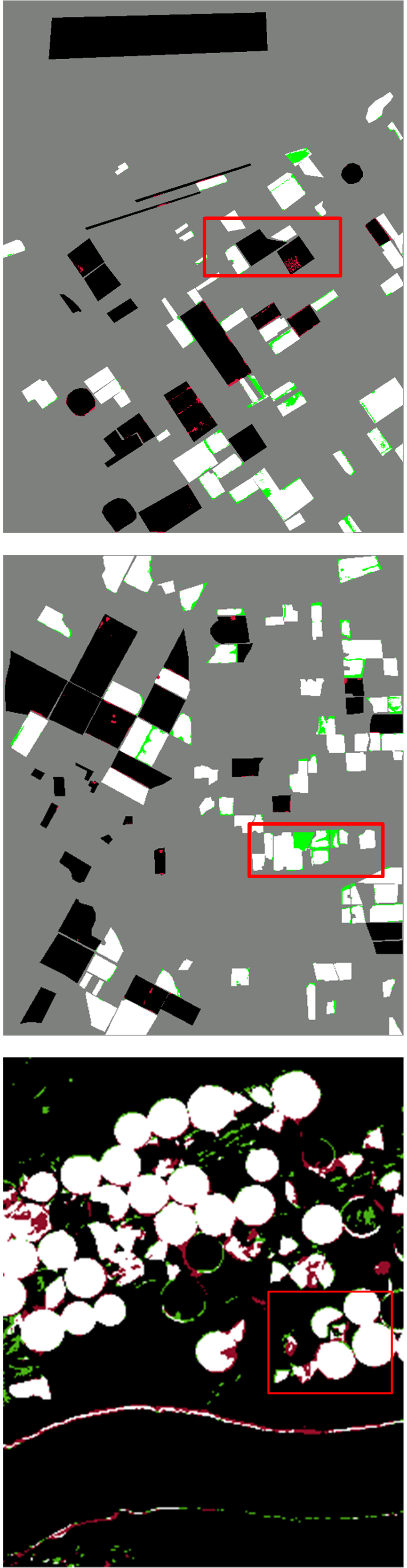}
    \end{minipage}
}
\subfigure[Ours]
{
 	\begin{minipage}[b]{.17\linewidth}
        \centering
        \includegraphics[scale=0.82]{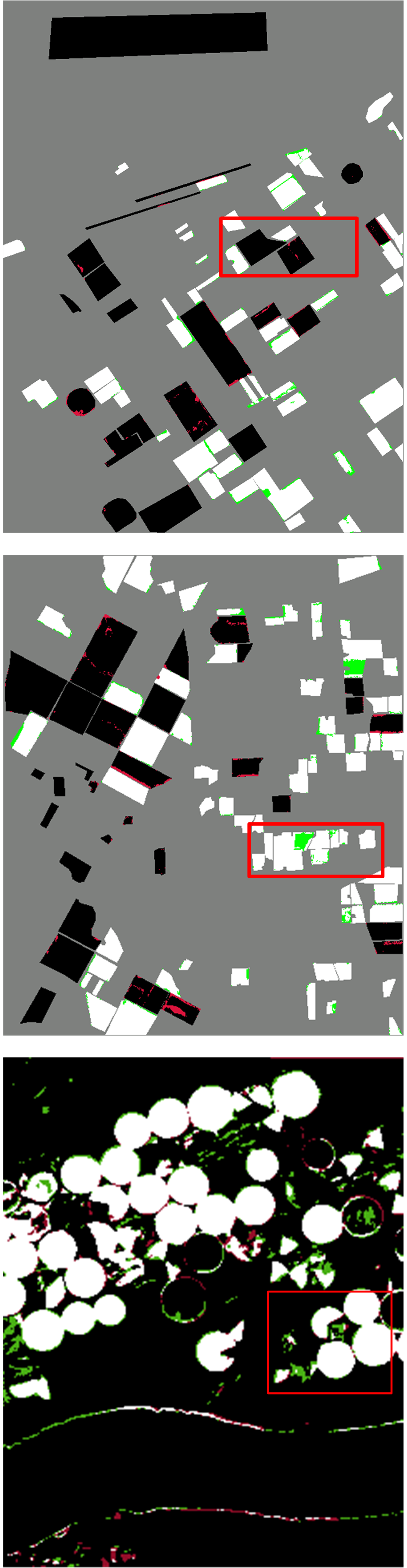}
    \end{minipage}
}
\subfigure[GT]
{
 	\begin{minipage}[b]{.17\linewidth}
        \centering
        \includegraphics[scale=0.82]{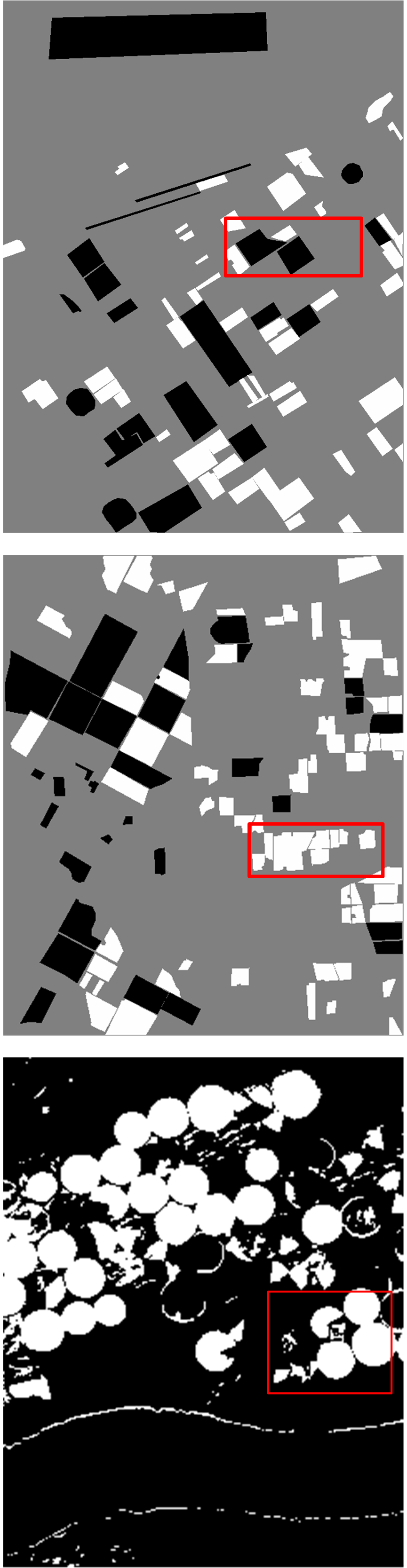}
    \end{minipage}
}%

\begin{minipage}[b]{0.07\textwidth}
    \centering
    \includegraphics[width=\textwidth]{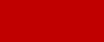}
\end{minipage}
\raisebox{\dimexpr\height--0\baselineskip}{\rotatebox[origin=c]{0}{\parbox[c][0.1cm][c]{1.2cm}{FP}}}
\begin{minipage}[b]{0.07\textwidth}
    \centering
    \includegraphics[width=\textwidth]{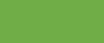}
\end{minipage}
\raisebox{\dimexpr\height--0\baselineskip}{\rotatebox[origin=c]{0}{\parbox[c][0.1cm][c]{1.2cm}{FN}}}
\begin{minipage}[b]{0.07\textwidth}
    \centering
    \includegraphics[width=\textwidth]{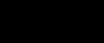}
\end{minipage}
\raisebox{\dimexpr\height--0\baselineskip}{\rotatebox[origin=c]{0}{\parbox[c][0.1cm][c]{1.2cm}{TN}}}
\begin{minipage}[b]{0.07\textwidth}
    \centering
    \includegraphics[width=\textwidth]{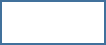}
\end{minipage}
\raisebox{\dimexpr\height--0\baselineskip}{\rotatebox[origin=c]{0}{\parbox[c][0.1cm][c]{1.2cm}{TP}}}
\begin{minipage}[b]{0.07\textwidth}
    \centering
    \includegraphics[width=\textwidth]{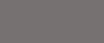}
\end{minipage}
\raisebox{\dimexpr\height--0\baselineskip}{\rotatebox[origin=c]{0}{\parbox[c][0.1cm][c]{3cm}{Unknown Area}}}

\caption{Visualizations of the proposed method and state-of-the-art unsupervised methods on three datasets. From top to bottom are Santa Barbara, Bay Area, and Hermiston datasets.}
\label{fig:barbara}
\end{figure*}

\subsubsection{Training the Cross-Temporal Contrastive Learning and Change Detection Head}

In the second stage, we keep the pre-trained SCDM parameters fixed and only focus on training the CTCL and CD head networks. Our goal is to learn features that are invariant to spectral differences caused by environmental changes. We use CTCL to align spectral feature representations of unchanged samples to achieve this. First, we obtain pseudo-labels using the traditional unsupervised method PCA \cite{deng2008pca} and then use them to train the entire network. We feed the original samples $X_{1}$ and $X_{2}$ into the CTCL to obtain contrastive feature representations $z_{i}$ and $z_{j}$. The loss function of the CTCL architecture based on the paper SimCLR \cite{chen2020simple} is defined as follows:

\begin{equation}
    \ell_{i, j}=-\log \frac{\exp \left(\operatorname{sim}\left(z_{i}, z_{j}\right) / \tau\right)}{\sum_{k=1}^{2 Q} \textbf{1}_{k \neq i} \cdot\left(\exp \left(\operatorname{sim}\left(z_{i}, z_{k}\right)\right) / \tau\right)}
    \label{eq:8}
\end{equation}
where $\textbf{1}_{[k \neq i]} \in\{0,1\}$ is an indicator function evaluating to1 if $k \neq i$.




\begin{equation}
    \mathcal{L}_{con}=\frac{1}{2 Q} \sum_{k=1}^{Q}[\ell(2 k-1,2 k)+\ell(2 k, 2 k-1)]
    \label{eq:9}
\end{equation}
where $\ell_{i, j}$ represents the loss of a pair of positive samples $(i, j)$, and $\mathcal{L}_{\text {con }}$ represents the total loss of contrastive learning. $\operatorname{sim}\left(z_{i}, z_{j}\right)$ is the cosine similarity between feature representations $z_{i}$ and $z_{j}$. $Q$ represents the number of unchanged samples in a sample set with a batch size of $N$. $\tau$ denotes a temperature parameter.

The CD task involves pixel-wise evaluation of changes at each location, and we use the cross-entropy loss to measure the change loss. The loss for variation is defined as follows:

\begin{equation}
    \mathcal{L}_{change}=-\frac{1}{N} \sum_{i=1}^{N}\left(y_{i} \log \hat{y}_{i}+\left(1-y_{i}\right) \log \left(1-\hat{y}_{i}\right)\right)
    \label{eq:10}
\end{equation}
where $y_{i} \in\{0,1\}$ represents the actual label, $0$ represents no change, $1$ represents a change, and $\hat{y}_{i}$ represents the label predicted by the network. Therefore, the total loss of our proposed DiffUCD framework is:

\begin{figure*}[htbp]
\centering
\subfigure[Base]
{
    \begin{minipage}[b]{.22\linewidth}
        \centering
        \includegraphics[scale=0.25]{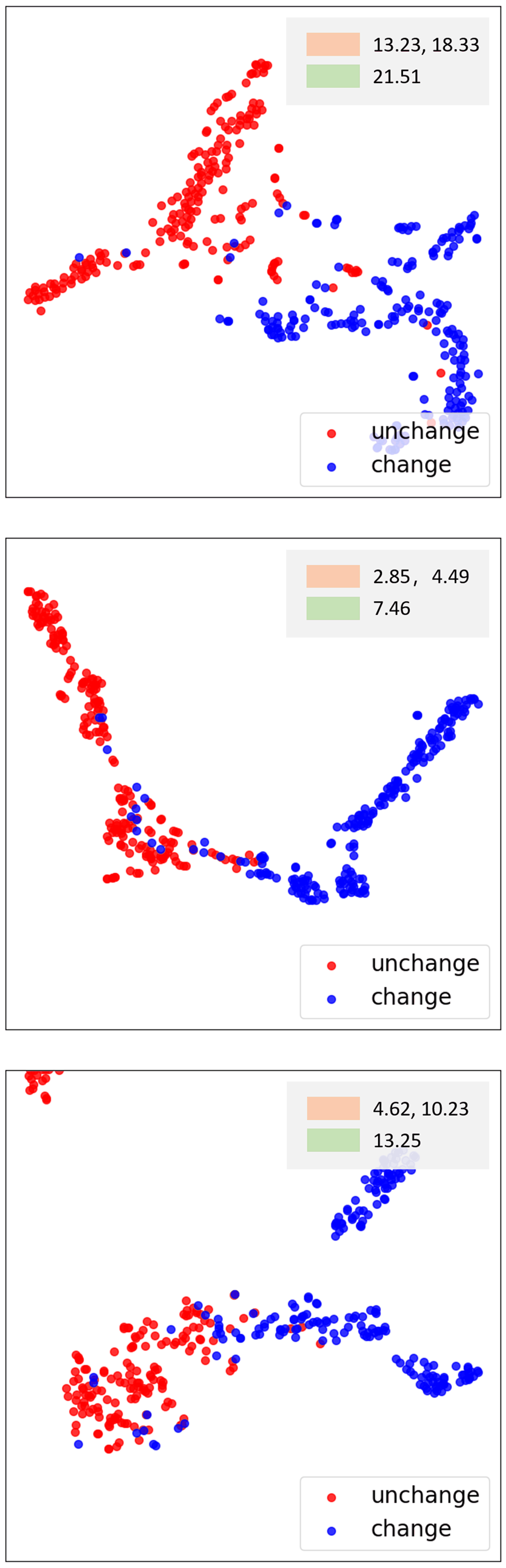}
    \end{minipage}
}%
\subfigure[Base+CTCL]
{
 	\begin{minipage}[b]{.22\linewidth}
        \centering
        \includegraphics[scale=0.25]{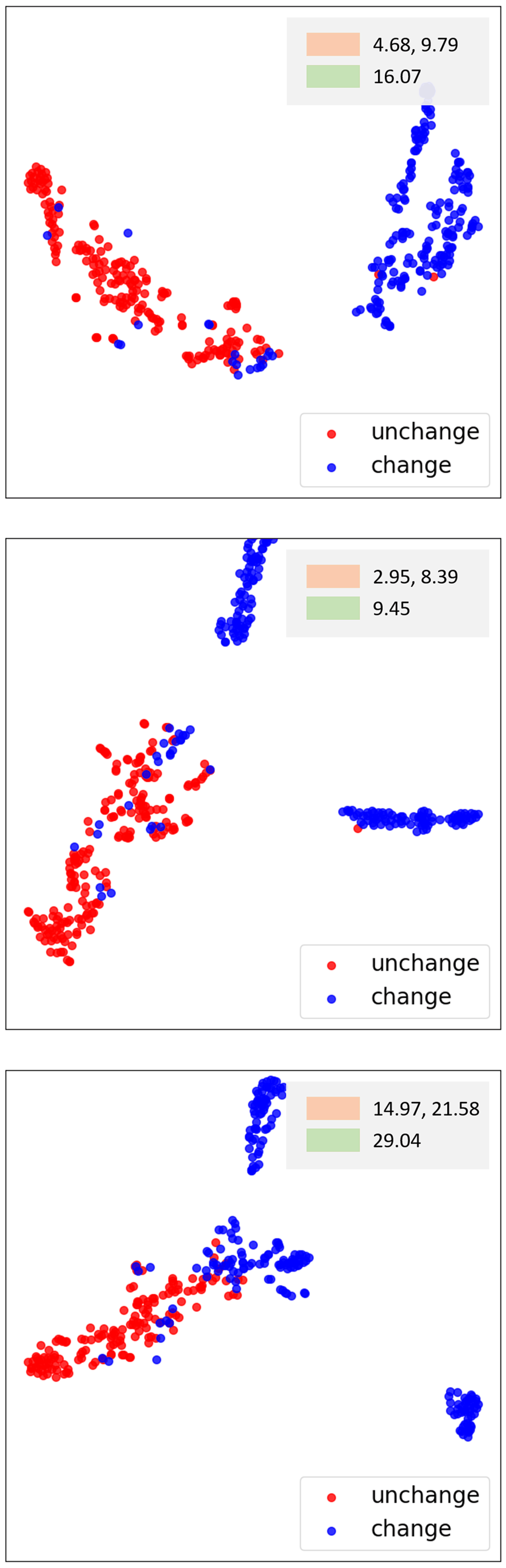}
    \end{minipage}
}
\subfigure[Base+SCDM]
{
 	\begin{minipage}[b]{.22\linewidth}
        \centering
        \includegraphics[scale=0.25]{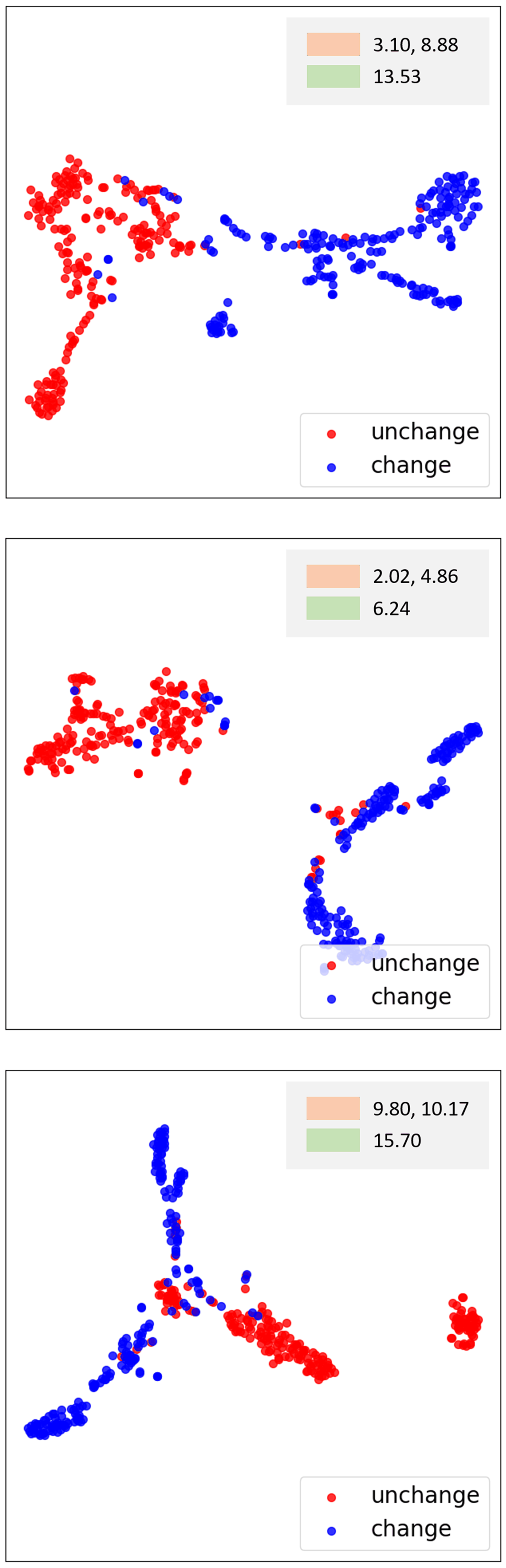}
    \end{minipage}
}
\subfigure[Ours]
{
 	\begin{minipage}[b]{.22\linewidth}
        \centering
        \includegraphics[scale=0.25]{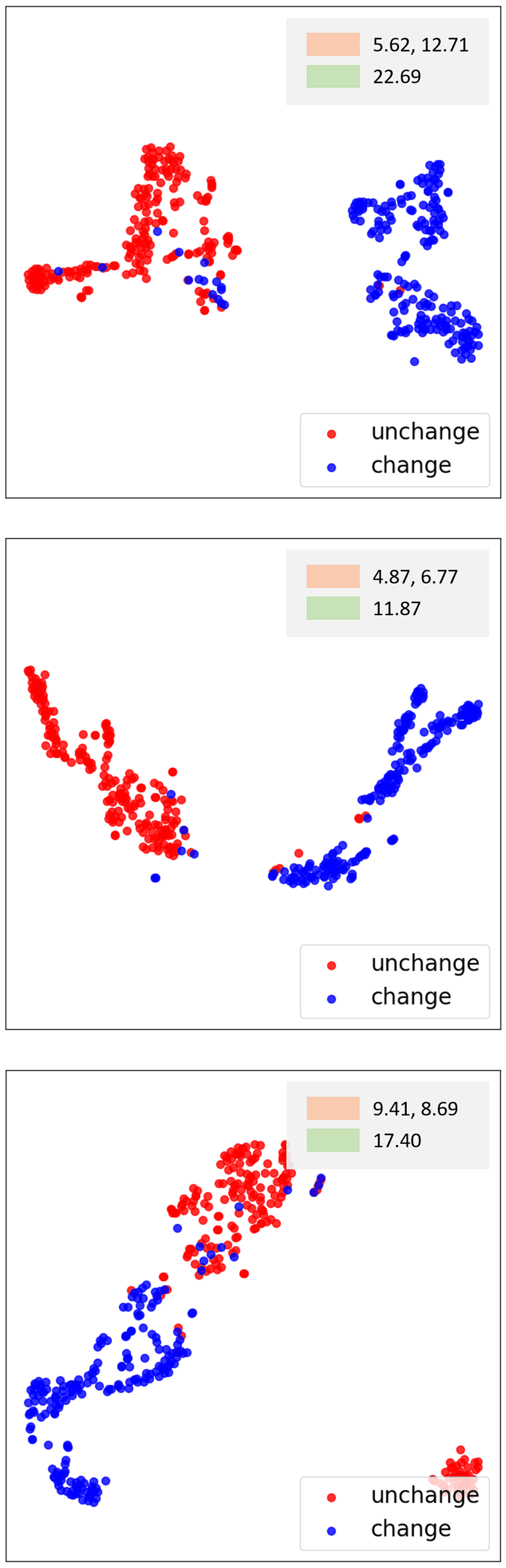}
    \end{minipage}
}

\begin{minipage}[b]{0.07\textwidth}
    \centering
    \includegraphics[width=\textwidth]{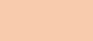}
\end{minipage}
\raisebox{\dimexpr\height--0\baselineskip}{\rotatebox[origin=c]{0}{\parbox[c][0.1cm][c]{3cm}{Intra-class Distances}}}
\begin{minipage}[b]{0.07\textwidth}
    \centering
    \includegraphics[width=\textwidth]{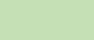}
\end{minipage}
\raisebox{\dimexpr\height--0\baselineskip}{\rotatebox[origin=c]{0}{\parbox[c][0.1cm][c]{3cm}{Inter-class Distances}}}

\caption{The t-SNE visualization of features extracted on three datasets. From top to bottom are Santa Barbara, Bay Area, and Hermiston datasets.}
\label{fig:tsne}
\end{figure*}

\section{Experiments}
\subsection{Datasets}

We demonstrate the effectiveness of our proposed method on three publicly available HSI-CD datasets: Santa Barbara, Bay Area, and Hermiston. The Santa Barbara dataset comprises imagery captured by the AVIRIS sensor over the Santa Barbara region in California. The dataset includes images from 2013 and 2014, with spatial dimensions of 984 $\times$ 740 pixels and 224 spectral bands. Similarly, the Bay Area dataset consists of AVIRIS sensor imagery surrounding the city of Patterson, California. The dataset includes images captured in 2013 and 2015, with spatial dimensions of 600 $\times$ 500 pixels and 224 spectral bands.

The Hermiston dataset focuses on an irrigated agricultural field in Hermiston, Umatilla County, Oregon. The imagery was acquired on May 1, 2004, and May 8, 2007. The image size is 307 $\times$ 241 pixels, consisting of 57,311 unchanged pixels and 16,676 changed pixels. After removing noise, 154 spectral bands were selected for the experiments. The changes observed in this dataset primarily pertain to land cover types and the presence of rivers.

\begin{table}[tbp]  
	\centering
	\caption{CONFUSION MATRIX
}
        \setlength{\tabcolsep}{5mm}{
	\begin{tabular}{c|c|cc} 			
		\toprule 
            \multicolumn{2}{c|}{\multirow{2}{*}{Confusion Matrix}} & \multicolumn{2}{c}{Predicted}\\
            \multicolumn{2}{c|}{}& Change & Unchange\\	
		\midrule 
		  \multirow{2}{*}{Actual} & Change & TP & FN\\
            & Unchange & FP & TN\\
		\specialrule{0em}{1pt}{1pt}
		\specialrule{0em}{1pt}{1pt}	
		\bottomrule  	
	\end{tabular}}
        \label{table:1}
\end{table}

Santa Barbara and Bay Area unlabeled pixels make up approximately 80$\%$ of all pixels. To train the CTCL and CD heads, we use the full-pixel pre-trained SCDM and select 500 changed and 500 unchanged pixels from the PCA-generated pseudo-labels \cite{deng2008pca}.

\begin{table}[htbp]
  \caption{COMPARISON WITH STATE-OF-THE-ART METHODS ON SANTA BARBARA DATASET}
  \label{table:comparison-santabarbara}
  \centering
  \setlength{\tabcolsep}{3.5mm}{
  \begin{tabular}{llll}
    \toprule
    \multicolumn{1}{c}{}    &   \multicolumn{3}{c}{Santa Barbara}  \\
    
    \cmidrule(r){2-4}
    
    Method & OA & KC & F1 \\
    \midrule
    CVA\cite{bovolo2006theoretical}& 87.12 & 73.10 & 83.78  \\
     PCA\cite{deng2008pca}& 88.40 & 76.76 & 86.95  \\
     ISFA \cite{wu2013slow}& 89.12 & 76.75 & 85.35\\
     DSFA \cite{8824216}& 87.70 & 73.23 & 82.49 \\
     MSCD \cite{9538396}& 78.68 & 53.13 & 68.72\\
     HyperNet \cite{hu2022hypernet}& 91.14 & 81.48 & 88.80\\
     \midrule
     Ours & \textbf{96.87} & \textbf{93.41} & \textbf{95.97}\\
     \midrule
     \multicolumn{4}{c}{\textbf{Supervised Model}}\\
     \midrule
     BCNNs \cite{lin2019multispectral} & 97.04 & 93.77 & 96.19 \\
     ML-EDAN \cite{qu2021multilevel}& 98.00 & 95.81 & 97.46\\
    \bottomrule
  \end{tabular}}
\end{table}

\begin{table}[htbp]
  \caption{COMPARISON WITH STATE-OF-THE-ART METHODS ON BAY AREA DATASET}
  \label{table:comparison-bayArea}
  \centering
  \setlength{\tabcolsep}{3.5mm}{
  \begin{tabular}{llll}
    \toprule
    \multicolumn{1}{c}{}  &   \multicolumn{3}{c}{Bay Area}\\
    
    \cmidrule(r){2-4}
    
    Method & OA & KC & F1 \\
    \midrule
    CVA\cite{bovolo2006theoretical}& 85.41 & 71.10  & 84.89 \\
     PCA\cite{deng2008pca}& 89.28 & 78.77  & 88.88 \\
     ISFA \cite{wu2013slow} & 89.17 & 78.48 & 89.05\\
     DSFA \cite{8824216}& 82.68 & 65.81 & 81.61  \\
     MSCD \cite{9538396} & 78.68 & 53.13 & 68.72\\
     HyperNet \cite{hu2022hypernet}& 90.79 & 81.52 & 91.29\\
     \midrule
     Ours & \textbf{96.35} & \textbf{92.67} & \textbf{96.57} \\
     \midrule
     \multicolumn{4}{c}{\textbf{Supervised Model}}\\
     \midrule
     BCNNs \cite{lin2019multispectral}& 96.84 & 93.67 & 96.97\\
     ML-EDAN \cite{qu2021multilevel}& 96.47 & 92.91 & 96.67\\
    \bottomrule
  \end{tabular}}
\end{table}

\begin{table}[htbp]
  \caption{COMPARISON WITH STATE-OF-THE-ART METHODS ON HERMISTON DATASET}
  \label{table:comparison-Hermiston}
  \centering
  \setlength{\tabcolsep}{3.5mm}{
  \begin{tabular}{llll}
    \toprule
    \multicolumn{1}{c}{}&   \multicolumn{3}{c}{Hermiston}\\
    
    \cmidrule(r){2-4}
   
    Method & OA & KC & F1 \\
    \midrule
    CVA\cite{bovolo2006theoretical}& 91.98 & 74.06 & 78.77   \\
     PCA\cite{deng2008pca}& 92.14 & 74.56 & 79.19   \\
     ISFA \cite{wu2013slow} & 90.23 & 67.16 & 72.62\\
     DSFA \cite{8824216} & 92.67 & 76.94 & 81.39  \\
     MSCD \cite{9538396} & 78.51 & 47.88 & 62.01\\
     HyperNet \cite{hu2022hypernet}& 92.06 & 76.13 & 81.12  \\
     BCG-Net \cite{hu2023binary}& 94.90 & 85.38 & 88.67\\
     \midrule
     Ours & \textbf{95.47} & \textbf{86.69} & \textbf{89.58}  \\
     \midrule
     \multicolumn{4}{c}{\textbf{Supervised Model}}\\
     \midrule
     BCNNs \cite{lin2019multispectral}& 93.39 & 81.49 & 85.79 \\
     \midrule
     ML-EDAN \cite{qu2021multilevel}& 94.58 & 84.89 & 88.41 \\
    \bottomrule
  \end{tabular}}
\end{table}

\begin{table*}[htbp]
  \caption{ABLATION EXPERIMENTS ON MODULE EFFECTIVENESS ON THREE DATASETS}
  \label{table:module}
  \centering
  \setlength{\tabcolsep}{3mm}{
  \begin{tabular}{llllllllllll}
    \toprule
    \multicolumn{3}{c}{}    &   \multicolumn{3}{c}{Santa Barbara}  &   \multicolumn{3}{c}{Bay Area}  &   \multicolumn{3}{c}{Hermiston}\\
    
    \cmidrule(r){4-6}
    \cmidrule(r){7-9}
    \cmidrule(r){10-12}
    Base     & SCDM     & CTCL & OA & KC & F1 & OA & KC & F1 & OA & KC & F1 \\
    \midrule
    $\sqrt{ }$ & & & 90.48 & 80.51 & 88.67 & 91.77 & 83.35  & 92.65 & 92.83 & 77.24 & 81.55   \\
    $\sqrt{ }$ & $\sqrt{ }$ & & 95.64 & 90.92 & 94.57 & 94.74 & 89.49 &  94.90 & 94.62 & 84.65 & 88.12   \\
    $\sqrt{ }$ & & $\sqrt{ }$ & 95.38 & 90.33 & 94.15 & 94.38 & 88.74 & 94.59 & 93.62 & 82.71 & 86.89\\
    $\sqrt{ }$ & $\sqrt{ }$ & $\sqrt{ }$ & \textbf{96.87} & \textbf{93.41} & \textbf{95.97} & \textbf{96.35} & \textbf{92.67} & \textbf{96.57} & \textbf{95.47} & \textbf{86.69} & \textbf{89.58}  \\
    \bottomrule
  \end{tabular}}
\end{table*}

\subsection{Experimental Details}
\subsubsection{Evaluation Metrics}
We quantitatively evaluate DiffUCD's performance using three widely-used metrics: Overall Accuracy (OA), Kappa Coefficient (KC), and F1 score. These metrics are used to comprehensively assess the model's accuracy, consistency, and balance between precision and recall. The above metrics are defined as follows:
\begin{equation}
    \mathrm{precision}=\frac{\mathrm{TP}}{\mathrm{TP}+\mathrm{FP}}
    \label{eq:pre}
\end{equation}


\begin{equation}
    \mathrm{recall}=\frac{\mathrm{TP}}{\mathrm{TP}+\mathrm{FN}}
    \label{eq:recall}
\end{equation}


\begin{equation}
    \mathrm{OA}=\frac{\mathrm{TP}+\mathrm{TN}}{\mathrm{TP}+\mathrm{TN}+\mathrm{FP}+\mathrm{FN}}
    \label{eq:OA}
\end{equation}

\begin{equation}
    \mathrm{KC}=\frac{\mathrm{OA}-\mathrm{PRE}}{1-\mathrm{PRE}}
    \label{eq:KC}
\end{equation}

\begin{equation}
    \mathrm{PRE}=\frac{(\mathrm{TP}+\mathrm{FP})(\mathrm{TP}+\mathrm{FN})}{(\mathrm{TP}+\mathrm{TN}+\mathrm{FP}+\mathrm{FN})^{2}}+\frac{(\mathrm{FN}+\mathrm{TN})(\mathrm{FP}+\mathrm{TN})}{(\mathrm{TP}+\mathrm{TN}+\mathrm{FP}+\mathrm{FN})^{2}}
    \label{eq:PRE}
\end{equation}

\begin{equation}
    \mathrm{F 1}=\frac{2}{\text { recall }^{-1}+\text { precision }^{-1}}
    \label{eq:F1}
\end{equation}

\subsubsection{Implementation Details}
We perform all experiments using the PyTorch platform, running on an NVIDIA GTX 2080Ti GPU with 11GB of memory. The batch size is 128, and a patch size of 7 is used to process the input data. In the first stage, the pre-training SCDM trains for 1000 epochs using the AdamW optimizer \cite{loshchilov2017decoupled} with an initial learning rate of 1e-5. The timestep for the SCDM was set to 200. In the second stage, we fix the parameters of the SCDM and use the Adadelta optimizer \cite{zeiler2012adadelta} to optimize the CTCL and CD head network over time. The initial learning rate is set to 1 and linearly decreases to 0 at 200 epochs. Through experiments, we choose the spectral-spatial features produced by the SCDM $t=5,10,100$ as the input features of the CD head.

\begin{figure*}[htbp]
\centering
\subfigure[t=200]
{
    \begin{minipage}[b]{.18\linewidth}
        \centering
        \includegraphics[scale=1.1]{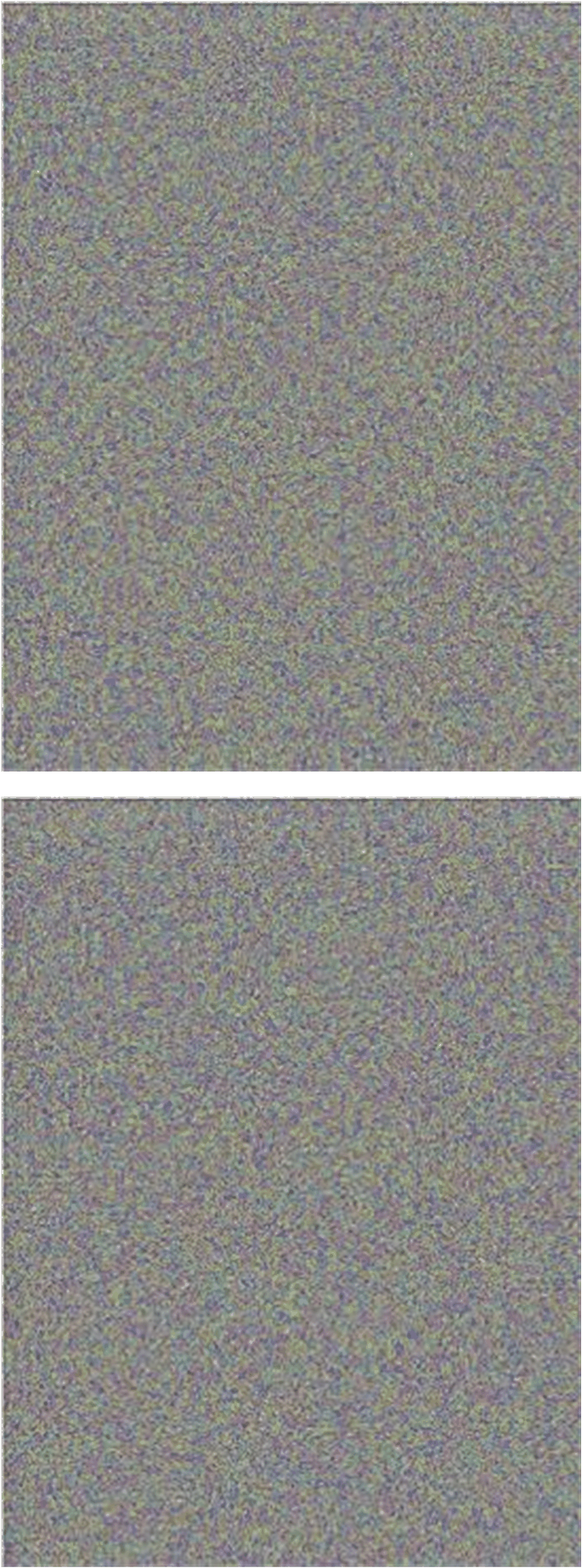}
    \end{minipage}
}%
\subfigure[t=50]
{
 	\begin{minipage}[b]{.18\linewidth}
        \centering
        \includegraphics[scale=1.1]{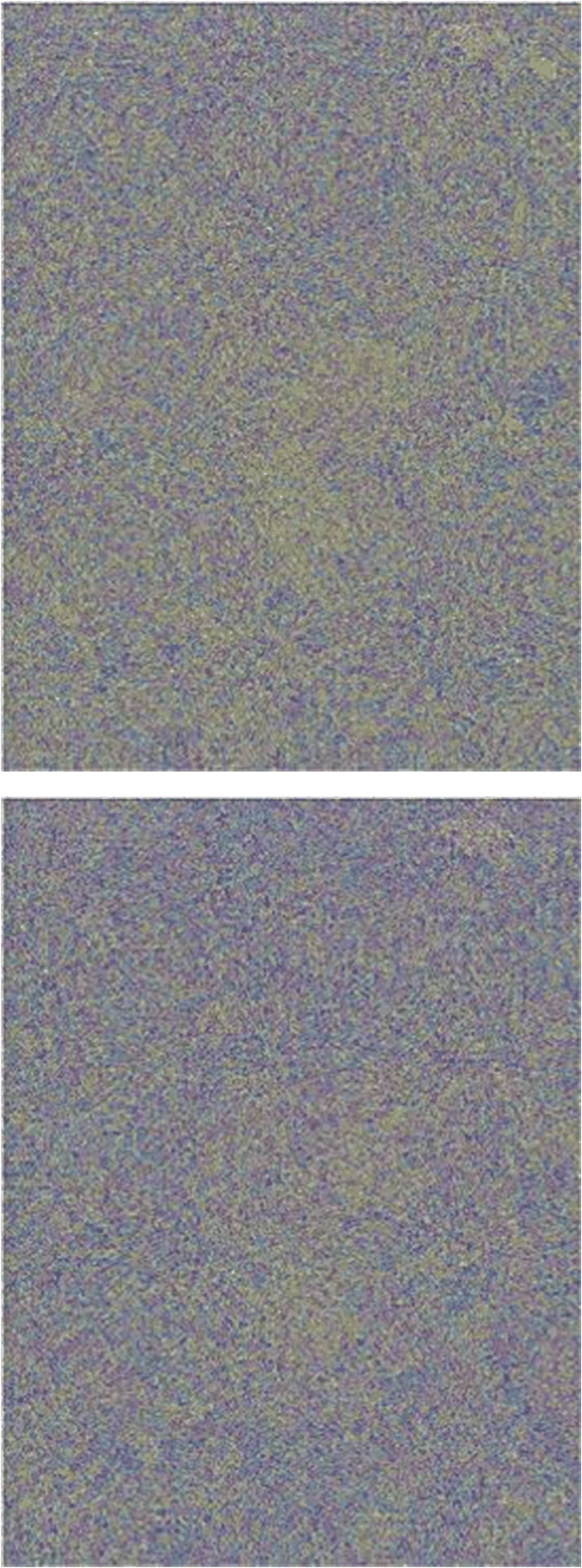}
    \end{minipage}
}
\subfigure[t=10]
{
 	\begin{minipage}[b]{.18\linewidth}
        \centering
        \includegraphics[scale=1.1]{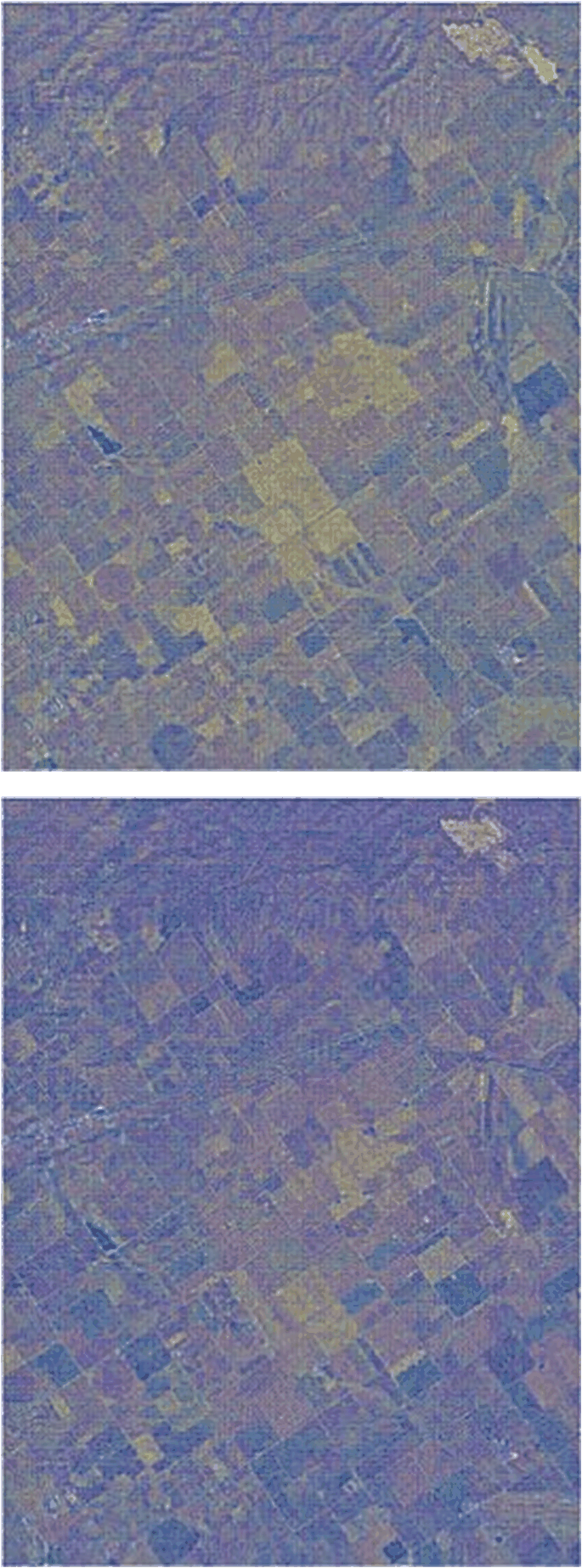}
    \end{minipage}
}
\subfigure[t=5]
{
 	\begin{minipage}[b]{.18\linewidth}
        \centering
        \includegraphics[scale=1.1]{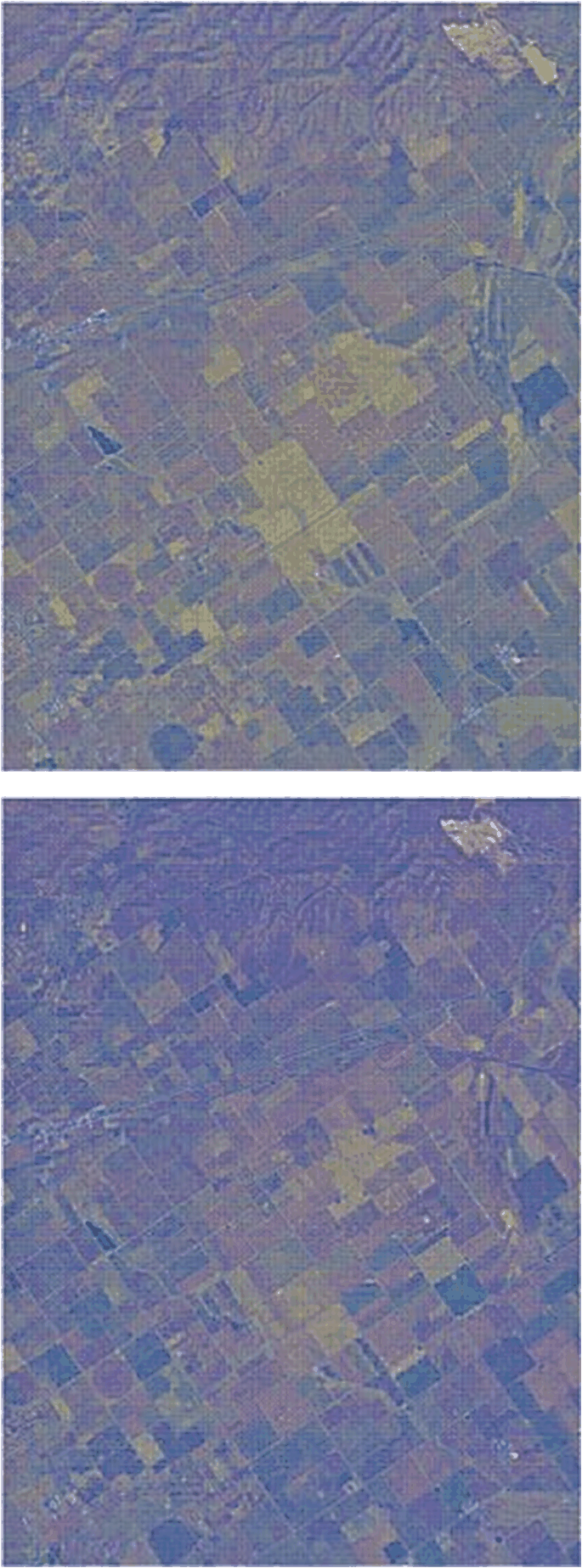}
    \end{minipage}
}
\subfigure[t=0]
{
 	\begin{minipage}[b]{.18\linewidth}
        \centering
        \includegraphics[scale=1.1]{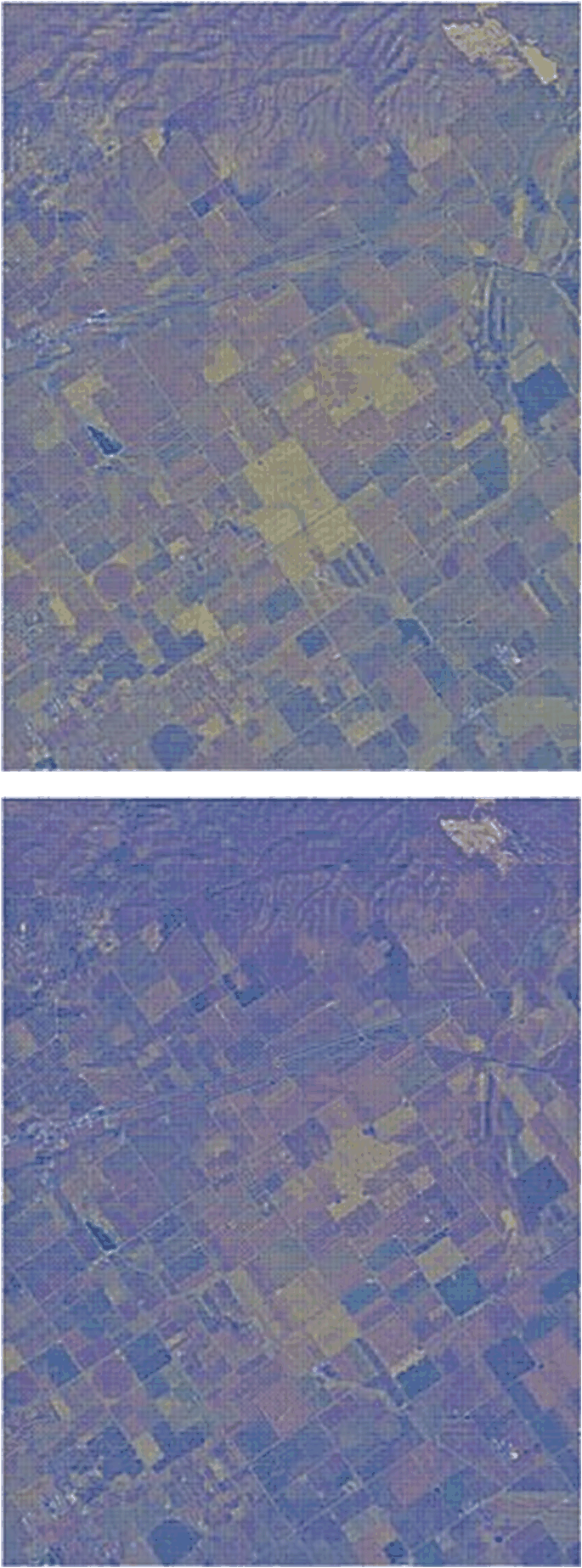}
    \end{minipage}
}

\caption{SCDM denoising process reconstructs pseudo-color images of different timestamps of the Santa Barbara dataset. Image visualization at time T1 and T2 from top to bottom.}
\label{fig:denoise_barbara}
\end{figure*}

\begin{figure*}[htbp]
\centering
\subfigure[t=200]
{
    \begin{minipage}[b]{.18\linewidth}
        \centering
        \includegraphics[scale=0.8]{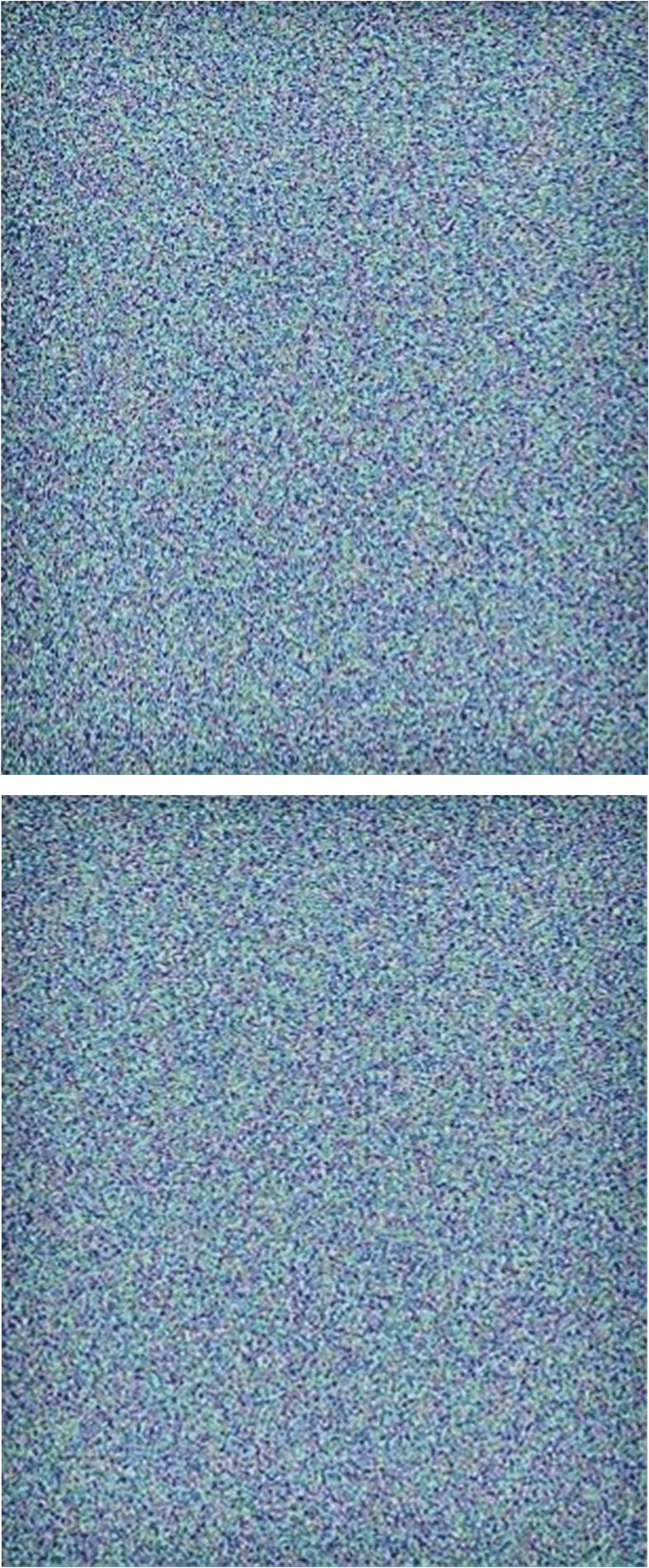}
    \end{minipage}
}%
\subfigure[t=50]
{
 	\begin{minipage}[b]{.18\linewidth}
        \centering
        \includegraphics[scale=0.8]{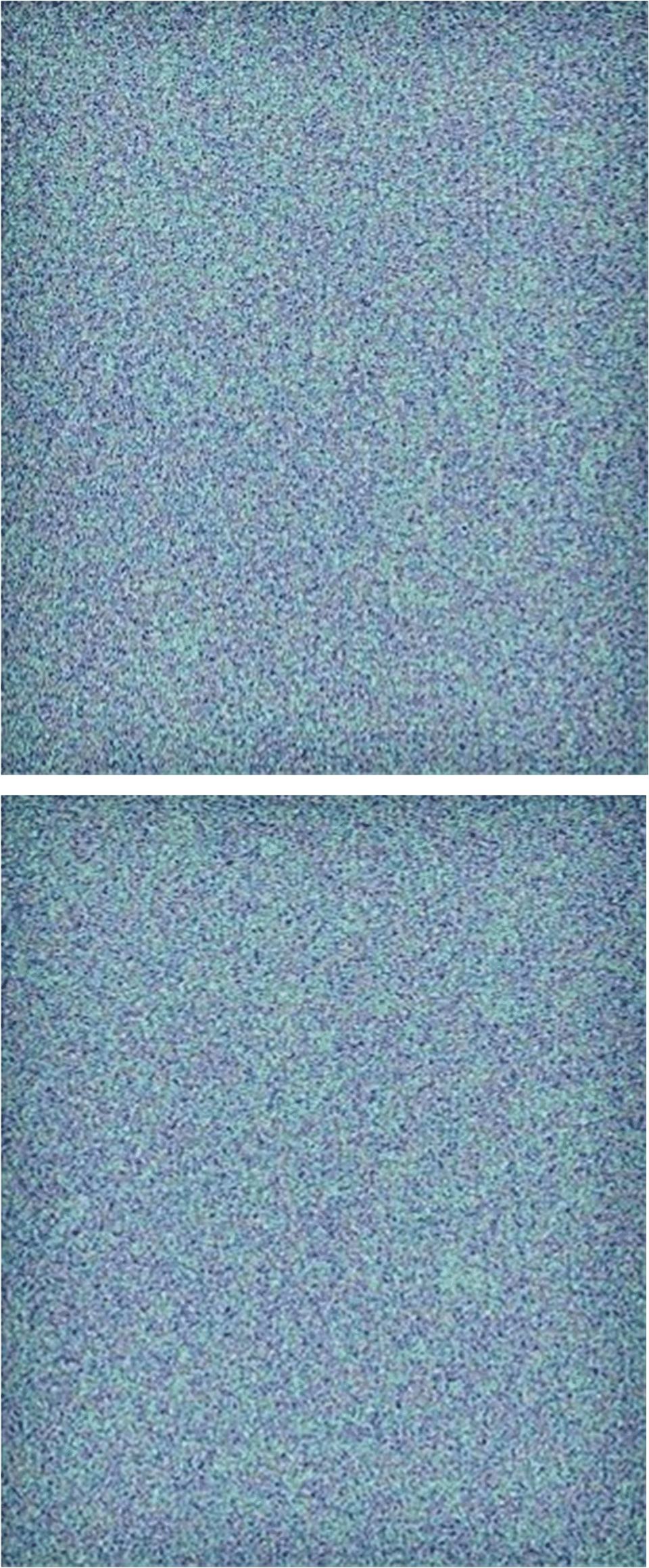}
    \end{minipage}
}
\subfigure[t=10]
{
 	\begin{minipage}[b]{.18\linewidth}
        \centering
        \includegraphics[scale=0.8]{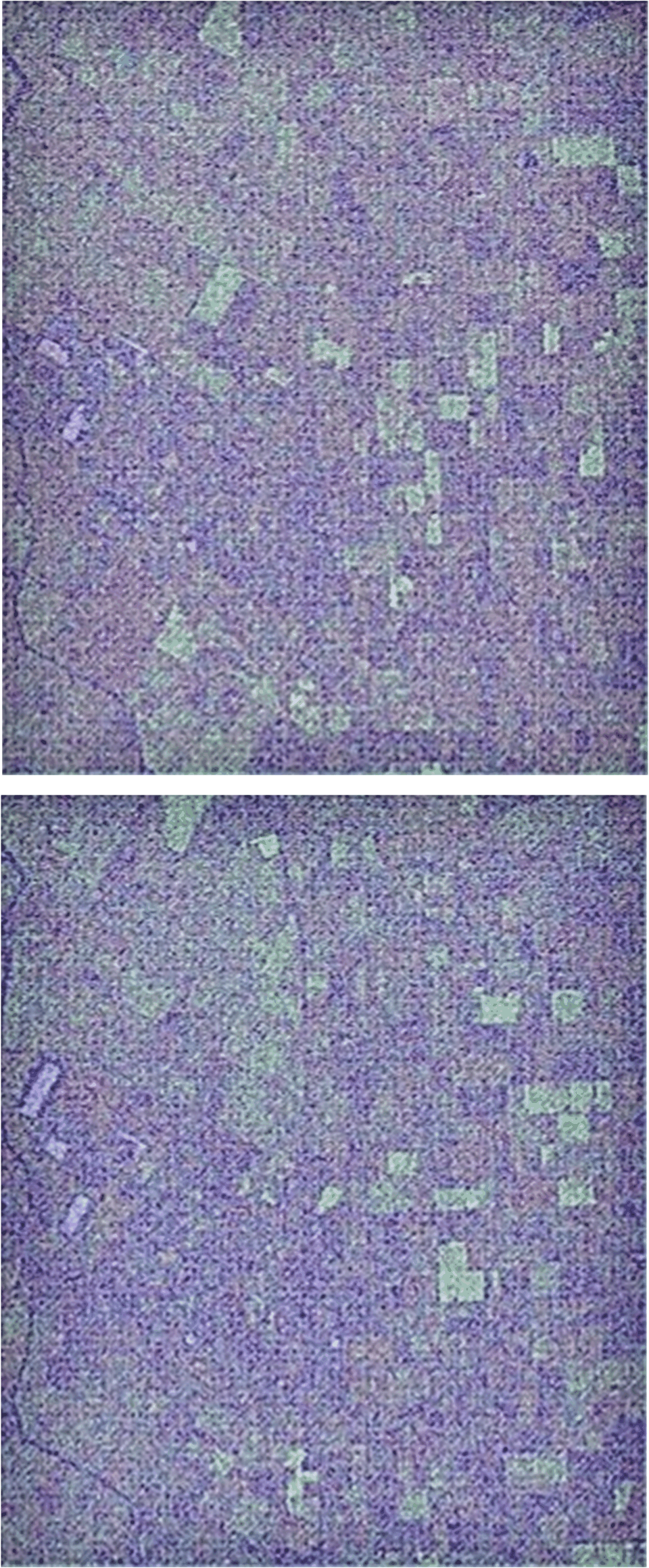}
    \end{minipage}
}
\subfigure[t=5]
{
 	\begin{minipage}[b]{.18\linewidth}
        \centering
        \includegraphics[scale=0.8]{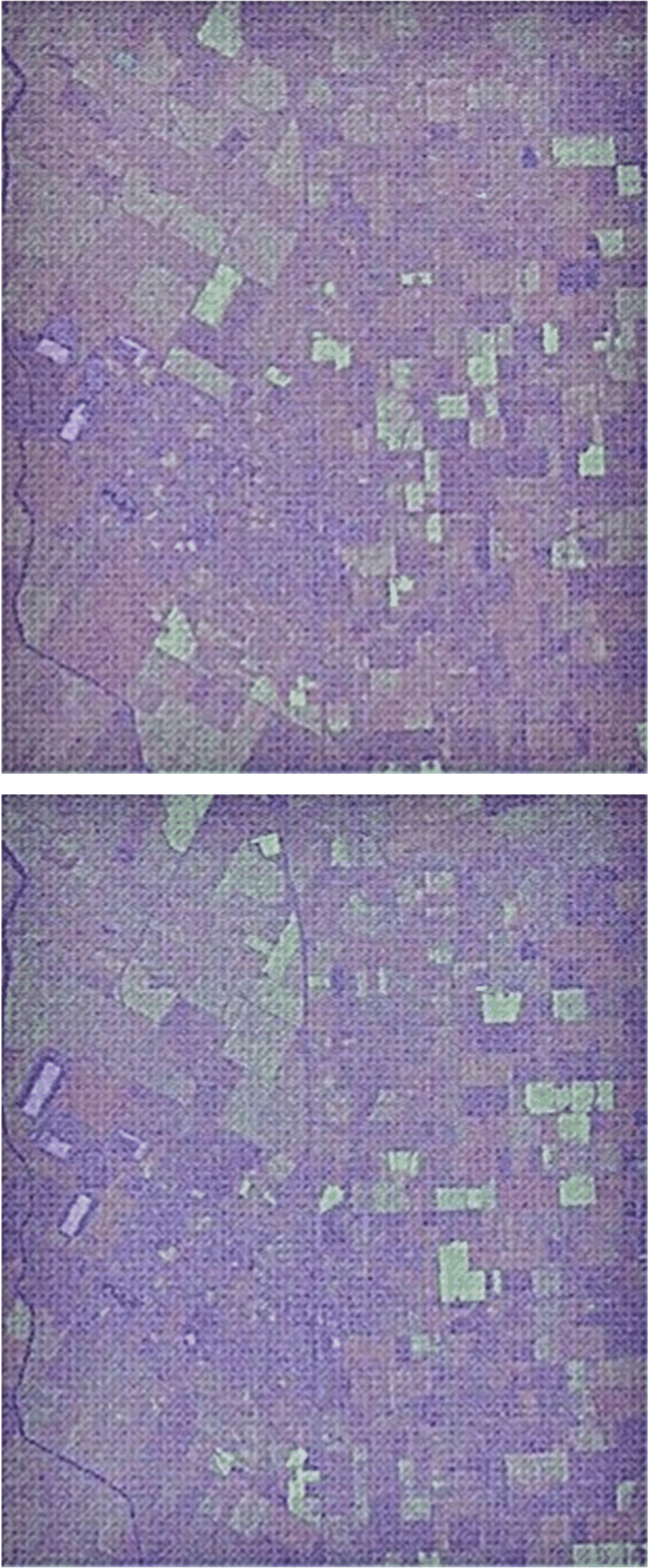}
    \end{minipage}
}
\subfigure[t=0]
{
 	\begin{minipage}[b]{.18\linewidth}
        \centering
        \includegraphics[scale=0.8]{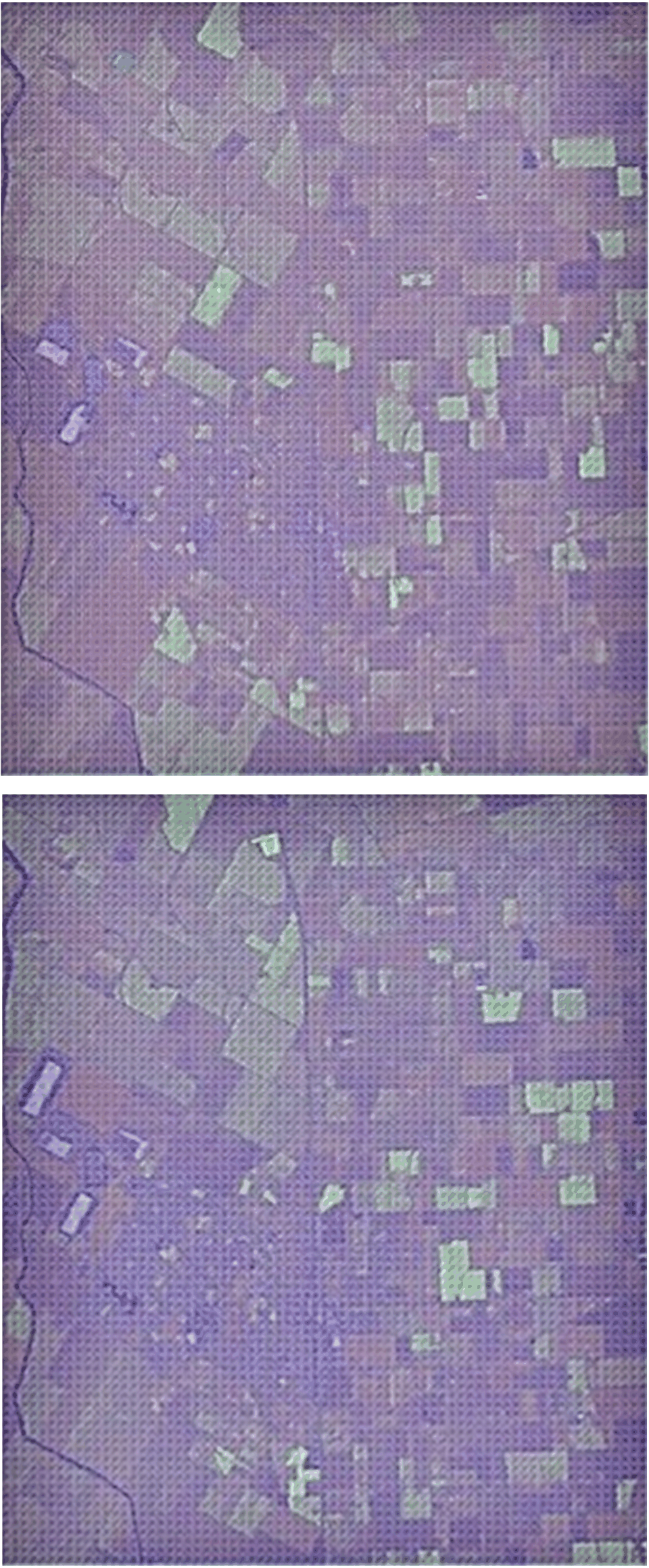}
    \end{minipage}
}

\caption{SCDM denoising process reconstructs pseudo-color images of different timestamps of the Bay Area dataset. Image visualization at time T1 and T2 from top to bottom.}
\label{fig:denoise_bayArea}
\end{figure*}

\subsection{Comparison to State-of-the-art Methods}

We conduct a comprehensive comparison of our method with recent unsupervised and supervised HSI-CD methods, including CVA \cite{bovolo2006theoretical}, PCA \cite{deng2008pca}, ISFA \cite{wu2013slow}, DSFA \cite{8824216}, MSCD \cite{9538396}, HyperNet \cite{hu2022hypernet}, BCG-Net \cite{hu2023binary}, BCNNs \cite{lin2019multispectral}, and ML-EDAN \cite{qu2021multilevel}. Fig. \ref{fig:barbara} presents a visual comparison of these methods on the three datasets.

From the visual observations in Fig. \ref{fig:barbara}, it is evident that our proposed method, DiffUCD, exhibits the smallest regions of red and green. This compelling visualization underscores the superior performance of DiffUCD compared to all other methods.
Table \ref{table:comparison-santabarbara}, Table \ref{table:comparison-bayArea}, and Table \ref{table:comparison-Hermiston} provides the quantitative results of DiffUCD alongside various state-of-the-art methods across the three datasets. Remarkably, our proposed method substantially improves performance over the state-of-the-art unsupervised methods, as evidenced by significant margins in OA, KC, and F1-score. Specifically, DiffUCD surpasses the unsupervised methods on the Santa Barbara dataset by remarkable margins of 5.73$\%$, 11.93$\%$, and 7.17$\%$ in terms of OA, KC, and F1-score, respectively. Furthermore, compared to supervised methods trained on an equivalent number of human-annotated training examples, our method demonstrates comparable or superior performance.

\subsection{Ablation Study}

\subsubsection{Effectiveness of the module}

We conduct a comprehensive ablation study to verify the effectiveness of the proposed SCDM and CTCL. The results are shown in Table \ref{table:module}. After adding the pre-training of the SCDM, the results of the network on the three datasets have been significantly improved. We argue that the SCDM pre-training process utilizes many unlabeled samples, which can extract the semantic correlation of spectral-spatial features of the CD dataset. The third row of Table \ref{table:module} is based on the base model, which adds a CTCL module, improving CD accuracy on the three datasets by aligning the spectral features of unchanged samples. The fourth row is the experimental results of the DiffUCD model we proposed, and the OA values on the three data sets have been increased by 6.39$\%$, 4.58$\%$, and 2.64$\%$, respectively. Experiments fully prove the effectiveness of our proposed DiffUCD and sub-modules.

\subsubsection{Comparison of feature extraction ability}

Fig. \ref{fig:tsne} visually demonstrates the effectiveness of the SCDM in extracting compact intra-class features compared to the base model. Notably, the feature distances obtained through the CTCL mechanism are significantly larger on the Santa Barbara and Hermiston datasets. The t-SNE visualization further reinforces the discriminative nature of our model. The t-SNE plot vividly illustrates that the features extracted by DiffUCD are well-separated, allowing for distinct clusters corresponding to different classes. This enhanced feature separability plays a crucial role in boosting CD accuracy.

\subsubsection{The influence of timestamp \textbf{$t$} on the reconstruction effect}
Fig. \ref{fig:denoise_barbara} and Fig. \ref{fig:denoise_bayArea} provides qualitative evidence of the effectiveness of DiffUCD in both noise removal and feature reconstruction of the original HSI. The visualization results clearly illustrate how the denoising process of DiffUCD fully incorporates the semantic correlation of spectral-spatial features, enabling the extraction of essential features that preserve the original image's semantic correlation.

\section{Conclusion}

This work presents a novel diffusion framework, called DiffUCD, designed explicitly for HSI-CD. To our knowledge, this is the first diffusion model developed for this particular task. DiffUCD leverages many unlabeled samples to fully consider the semantic correlation of spectral-spatial features and retrieve the features of the original image semantic correlation. Additionally, we employ CTCL to align the spectral feature representations of unchanged samples. This alignment facilitates learning invariant spectral difference features essential for capturing environmental changes. We evaluate the performance of our proposed method on three publicly available datasets and demonstrate that it achieves significant improvements over state-of-the-art unsupervised methods in terms of OA, KC, and F1 metrics. Furthermore, the diffusion model holds great potential as a novel solution for the HSI-CD task. Our work will inspire the development of new approaches and foster advancements in this field.



\bibliographystyle{IEEEtran}
\bibliography{IEEEabrv,mybibfile}
\ifCLASSOPTIONcaptionsoff

\newpage
\fi
\newpage
\newpage
\newpage

%

\end{document}